# EFFICIENT PREDICTION OF DNA-BINDING PROTEINS USING MACHINE LEARNING

S. Qatawneh[1], A. Alneaimi[1], Th. Rawashdeh[1], M. Muhairat[1], R. Qahwaji[2] and S. Ipson[2]

[1]Department of Software Engineering, Faculty of Science and Information Technology,
Al-Zaytoonah University of Jordan, Amman, Jordan
`S.Qatawneh@zuj.edu.jo`

[2] Informatics Research Institute, School of Computing, Informatics & Media, University
of Bradford, Richmond Road, Bradford, West Yorkshire, BD7 1DP, UK
`S.S.Ipson@Bradford.ac.uk`

## ABSTRACT

*DNA-binding proteins are a class of proteins which have a specific or general affinity to DNA and include three important components: transcription factors; nucleases, and histones. DNA-binding proteins also perform important roles in many types of cellular activities. In this paper we describe machine learning systems for the prediction of DNA- binding proteins where a Support Vector Machine and a Cascade Correlation Neural Network are optimized and then compared to determine the learning algorithm that achieves the best prediction performance. The information used for classification is derived from characteristics that include overall charge, patch size and amino acids composition. In total 121 DNA- binding proteins and 238 non-binding proteins are used to build and evaluate the system. For SVM using the ANOVA Kernel with Jack-knife evaluation, an accuracy of 86.7% has been achieved with 91.1% for sensitivity and 85.3% for specificity. For CCNN optimized over the entire dataset with Jack knife evaluation we report an accuracy of 75.4%, while the values of specificity and sensitivity achieved were 72.3% and 82.6%, respectively.*

## 1. INTRODUCTION

DNA-binding proteins (DNA-BPs) represent a broad category of proteins. They are divided into 54 protein-structural families according to their structures [1]. DNA-binding proteins are involved in many areas including the organization of genetic expression (transcription factors) [2], and in the packaging of DNA inside the nucleus (histones) [3]. These binding proteins consist of the zinc finger, the helix turn helix, the leucine zipper domain, the winged helix, the ETS domain, the Helix-loop-helix and the unusual DNA-binding domain consisting of the Immunoglobulin Fold [4][5].

The interactions between protein and DNA control several cellular activities, such as transcription, replication, DNA repair, recombination and other critical steps in cellular development [1]. Consequentially, these interactions have been of considerable interest to molecular biologists over the past three decades, and more recently, to computational biologists [9][10].

  1

Previously, a number of computational methods have been developed for automated identification and prediction of DNA-BPs based on the features derived from sequence and structure collectively and those derived from sequence alone. Ahmad et al. [11] developed

SeqPredNet neural networks using sequential features and achieved 40.6% sensitivity and 76.2% specificity with 3-fold cross validation, which is similar to holdout evaluation. They further integrated overall charge and electric moment into the identification of DNA-BPs, and achieved a higher performance with 68.2% sensitivity and 66% specificity, with 6-fold cross-validation [12]. Jones et al. [13] developed a prediction method to characterize electrostatic features of proteins for an automated approach to DNA-BP and DNA-binding site prediction. Correct predictions were made for 68% of the data set. Using the same dataset, Kuznetsov et al. [14] employed a Support Vector Machine for the identification of DNA-binding sites on the basis of sequential/structural features with 71% accuracy and balanced values of sensitivity and specificity. Yan et al. [15] developed a Naïve Bayes classifier to predict whether a given amino acid residue is a DNA-binding residue based on its identity and the identities of its sequence neighbours and achieved 78% accuracy. Bhardwaj et al. [16] implemented an SVM approach and reported 100% accuracy for self-consistency technique. Using "leave 1-pair holdout" evaluation, an accuracy of 86.3% was achieved. They also reported an accuracy of 90% for N-fold cross-validation, implemented using two methods: 5-fold cross-validation and one left out method. Despite the recent advances in this field, as highlighted above, some problems still exist. In [16], the authors achieved 100% accuracy for self-consistency when training and testing an SVM using the same database. However, this is not a good measure of SVM efficiency because when we build a model we want to use it to classify new data. Hence we are chiefly interested in model performance on new (unseen) data [52]. So we should use two data set: the training set (seen data) to build the model (determine its parameters) and the test set (unseen data) to measure its performance (holding the parameters constant). Both data sets have to be representative samples of the data that the model will be applied to [52].

Using the "one left out" N-fold cross-validation method, which leaves each protein out for testing exactly once, Bhardwaj et al. reported 90% accuracy. In this technique, the whole proteins are randomly separated into N parts. Each time, one part is retained for testing and all others form training dataset. This is repeated until each part forms the testing dataset exactly once [16]. It should be noted that they called this method Jack-knife testing but it is not the standard Jack-knife method normally used in machine learning work.

In this work, we aim to tackle some of these problems by using the Jack-knife technique which is widely recognised as an appropriate and efficient validation method, as explained later. In this paper, we describe the development of machine learning algorithms which distinguish DNA-BPs from non-binding ones with high accuracy and the investigation of different features depending on the structures and sequences of DNA-BPs. Two efficient machine learning algorithms: CCNN and SVM are optimised and then used in our prediction experiments.

The organization of the paper is as follows. In the second section, the Dataset and the discriminative powers of individual features that are used in this work are described. The implementation of Machine learning Algorithms (SVM and CCNN) is presented in the third section. In the fourth section, the evaluation method adopted is discussed together with the experimental results and performance. Concluding remarks and suggestions for possible future work are given in the last section.

## 2. METHODOLOGY AND SYSTEM DESIGN

The following formatting rules must be followed strictly. This (.doc) document may be used as a template for papers prepared using Microsoft Word. Papers not conforming to these requirements may not be published in the conference proceedings.

This section presents a brief description of the data used, the feature extraction techniques employed and how these features are implemented for use with Machine learning algorithms. Figure 1 shows in diagrammatic form the elements of the system design.

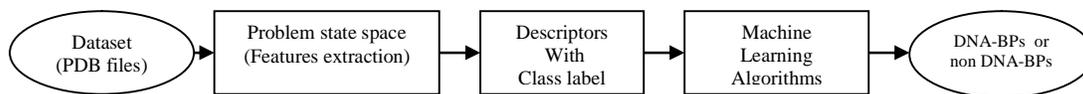

Figure1. Block diagram representing our prediction system

### 2.1. Dataset

The work described in this paper, used a positive dataset of 121 DNA-BPs [11-14] and a negative dataset of 238 non-DNA-BPs [13][15][16] obtained from a union of datasets previously used in related studies. These data are maintained by the Research Collaborators of Structural Bioinformatics (RCSB) (http://www.rcsb.org/pdb). Each entry in the PDB is identified by a unique 4-letter code, beginning with a number and followed by three alpha-numeric characters such as 1SHA. If the structure with the PDB code chosen has multiple chains then it is also necessary to specify the chain identifier (a single alpha-numeric character). Otherwise the field is blank. A complete list of the proteins used here is available at (http://proteomics.bioengr.uic.edu/pro-dna).

The structures of the complexes in both positive and negative cases were measured with better than 3Å resolution. These proteins have < 35% sequence identity between each pair [16][17].

### 2.2. Feature Extraction

The system attempts to successfully predict if a given protein belongs to the DNA-BP or non-DNA-BP classes. The proteins are translated into a state space for the problem which contains some characteristics representing features that may distinguish between the two classes. Also, every member in the dataset is given one of the two class labels: DNA-binding or non-DNA-binding for system testing and evaluation.

The features studied in this paper, obtained from previous research [16], include overall charge of the protein, patch size, overall amino acid composition and amino acid surface composition. Each of these features is described in the following section.

#### 2.2.1. Overall charge

The overall charge of a protein represents the strongest attribute used in the machine learning system. However, x-ray crystallography cannot resolve the presence of hydrogen atoms in most protein crystals, which means that in most PDB files, hydrogen atoms are absent [18]. Therefore, hydrogen atoms are added to all the measured protein structures using the publicly available tool REDUCE [19]. Then the form of the potential energy function is taken directly from CHARMM force field parameters to obtain charges for all the atoms [20].

As shown in Figure 2, the distributions of overall charge show significant differences for binding and non-binding proteins. More than 65% of binding proteins were positively charged whereas only 15% of the binding proteins were positively charged, so the overall, charge can be expected to distinguish between the two cases with a limited accuracy.

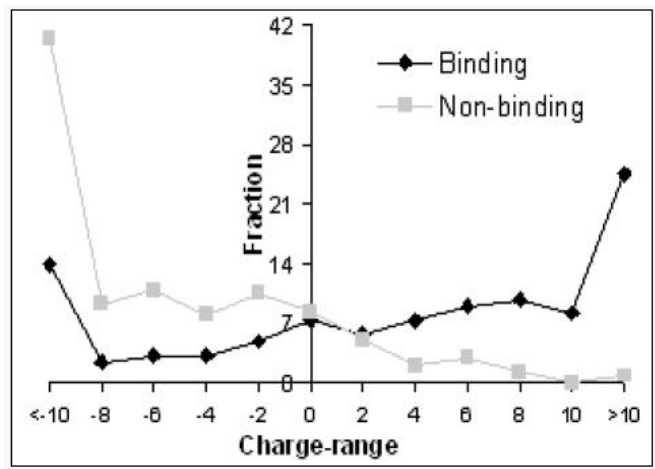

Figure 2. Distribution of overall charge for binding and non-binding cases in units of electron charge (e). Labels on the horizontal axis indicate the upper value of the binding [16].

### 2.2.2. Electrostatic calculations and patch size

To extract this feature the electrostatic potential is computed using the software package Delphi [16][17][21-25]. The potential is computed for individual protein atoms, with the DNA removed. This tool solves the non-linear Poisson–Boltzmann equation using finite-difference methods to calculate the potentials at specified points. Debye-Huckel boundary conditions were employed (the default for this package) and a simplified charge set defined from the molecular dynamics package CHARMM22 [26]. Additional processes, using the DSSP program [27], calculate the exposed surface area of all residues. A residue is classified as a "surface" residue if it has more than 40% of its total area exposed to water; otherwise it is classified as "buried" [17]. Furthermore, a surface residue is defined as "binding" if one of its heavy atoms is within a distance of 4.5 Å of any atom of the DNA [17].

The size of a patch is defined by the number of atoms it contains. Usually more then one patch is formed on each protein [16] and the size of the largest patch is used as a feature in the machine learning system.

### 2.2.3. Amino acid composition

Overall and surface amino acid compositions are computed for a protein, each kind representing 20 descriptors (percentages of the 20 amino acids). Figure 3 illustrates the frequency of occurrence of different amino acids in both cases.

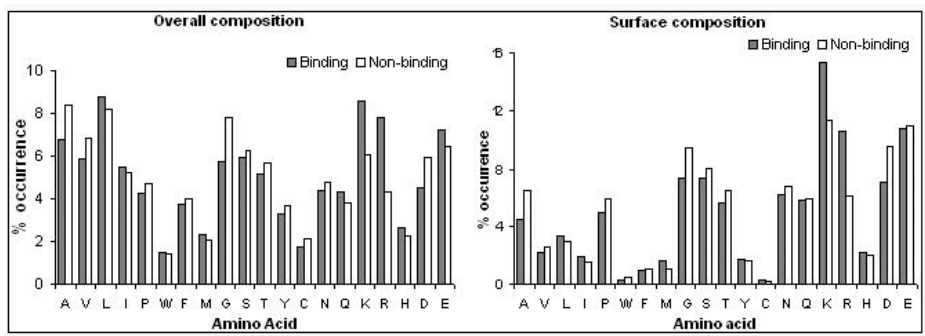

Figure 3. The frequency of occurrence of different amino acids is shown for overall and surface compositions. The difference in the height of a bar for an amino acid for binding and non-binding cases reflects a stronger preference for that amino acid in one case over the other [16].

### a. Overall composition

Here the whole protein is considered; because a positive charge is one of the common properties of DNA-binding proteins [28] and Lys and Arg amino acids are both positively charged their over-representation in DNA-BPs is fairly clear [16][28].

### b. Surface composition

When the surface of the protein is considered, typically only 40% of the whole surface is used. Surface composition is more discriminating than overall composition for hydrophobic residues such as Trp, Phe, Tyr, Cys, Ile and Met [16][17]. Binding and non-binding proteins have similar amino acid compositions [16].

Because of the limited discriminatory properties of the above compositions, this work like others [16][17], has used them in conjunction with other features.

## 3. MACHINE LEARNING ALGORITHMS

One of the latest subjects of discussion by machine learning researchers is the comparative performance of Support Vector Machine (SVM) and Neural Networks (NN). The work described in this paper compares SVM and Cascade Correlation Neural Network (CCNN) in terms of correct DNA-BPs prediction. CCNN were chosen over other NN architectures because of their excellent performance in classification [25]. Work reported by researches in applications [26][27] not related to the DNA-BP prediction, has found SVMs to out perform NNs in some classification tasks.

### 3.1. Support Vector Machine

Support Vector Machines are becoming popular tools for solving a variety of learning and function estimation problems. In contrast to neural networks, SVMs have a firm statistical foundation and are guaranteed to converge to the global minimum during training. They are also considered to have better generalization capabilities than neural networks and have managed to outperform neural networks in a number of applications [26-31]. SVMs employ a statistical learning theory based algorithm, introduced by V.Vapink [32], which uses the structural risk minimization principle. SVMs use non-linear-transformation kernel functions to map the input data to a high dimensional feature space, where data can become linearly separable [27][32].

In the work described here, the mySVM program developed by S.Ruping [33] was used to build the classifier. This program is based on the optimisation algorithm SVM light [34], which is equivalent to solving the following optimization problem:

Maximize

$$W(\alpha, \alpha^*) = -\frac{1}{2} \sum_{i,j=1}^{l} (\alpha_i^* - \alpha_i)(\alpha_j^* - \alpha_j)(x_i \cdot x_j) + \sum_{i=1}^{l} y_i(\alpha_i^* - \alpha_i) - \sum_{i=1}^{l} \varepsilon(\alpha_i^* - \alpha_i) \quad (1)$$

With respect to

$$0 \leq \alpha_i, \alpha_i^* \leq C \quad , i = 1, ..., n \quad (2)$$

$$\sum_{i=1}^{l} \alpha_i^* = \sum_{i=1}^{l} \alpha_i \quad (3)$$

Where (xi, yi) are a set of examples, C is a capacity constant and    ,   are variables specific to each example.

### 3.2. Cascade Correlation Neural Network

The training of back-propagation neural networks is considered to be a slow process because of the step-size and moving target problems [35]. To overcome these problems cascade neural networks were developed. These are "self organizing" networks [35] with topologies which are not fixed. The supervised training begins with a minimal network topology and new hidden nodes are incrementally added to create a multi-layer construction. The new hidden nodes are added to make the most of the correlation between the new node's output and the remaining error signal that the system is being adjusted to eliminate. The weights of a new hidden node is fixed and not changed later, hence making it a permanent feature detector in the network. This feature detector can then be used to generate outputs or to create other more complex feature detectors [35].

In a CCNN, the number of input nodes is determined by the input features, while the number of output nodes is determined by the number of different output classes. The training of a CCNN starts with no hidden nodes. The direct input-output connections are trained using the entire training set with the aid of the back propagation learning algorithm. Hidden nodes are then added gradually and every new node is connected to every input node and to every pre-existing hidden node.

Training is carried out using the training vector and the weights of the new hidden nodes are adjusted after each pass [35]. Cascade correlation networks have a number of attractive features including a very fast training time, often a hundred times faster than a perceptron network [35]. This makes cascade correlation network suitable for use with large training sets. Typically, cascade correlation networks are fairly small, often having fewer than a dozen neurons in the hidden layer [36][37]. This can be contrasted with probabilistic neural networks which require a hidden-layer neuron for each training case. Also, the training of CCNNs is quite robust, and good results can usually be obtained with little or no adjustment of parameters [35-37].

As with all types of systems, there are some potential pitfalls to using CCNNs. They have an extreme potential for over fitting the training data [38]; this results in excellent accuracy on the training data but poor accuracy on new, unseen data [39][40]. CCNNs are usually less accurate than probabilistic and general regression neural networks on small to medium size problems but scale up to handle large problems far better than probabilistic or general regression networks [35].

## 4. DISCUSSION

The values for negative and positive results in an experiment are likely to be the most useful practically. These values are influenced by the various performance criteria. In order for consistency with all other references [8-14] the performance criteria used are accuracy (the fraction of all correct predictions), sensitivity (the fraction of positive cases correctly classified) and specificity (the fraction of negative cases correctly classified). These are defined as follows:

$$Accuracy = \frac{TP+TN}{TP+TN+FP+FN} \quad (4)$$

$$Specificity = \frac{TN}{TN+FP} \quad (5)$$

$$Sensitivity = \frac{TP}{TP+FN} \quad (6)$$

The numbers of true positive, true negative, false positive and false negative results are indicated by TP, TN, FP and FN respectively, defined as follows:

$$TP = \frac{Correct\ positive\ predictions}{Total\ positives} \quad (7)$$

$$FP = \frac{Incorrect\ negative\ predictions}{Total\ negatives} \quad (8)$$

$$TN = \frac{Correct\ negative\ predictions}{Total\ negative} \quad (9)$$

$$FN = \frac{Incorrect\ positive\ predictions}{Total\ positive} \quad (10)$$

In these equations: "Correct positive predictions" is the total number of cases for which the system correctly predicts a binding protein; "Incorrect positive predictions" is the total number of cases for which the system incorrectly predicted that a protein is binding; "Correct negative predictions" is the total number of cases for which the system correctly predicts a non binding protein; "Incorrect negative predictions" is the total number of cases for which the system incorrectly predicted that a protein is non binding; "Total positives" is the sum of cases for binding proteins (Number of associated cases used in testing); "Total negatives" is the sum of cases for which a protein is non binding (Number of un-associated cases used in testing) [45][46].

### 4.1 Jack-knife Technique

The Jack-knife evaluation technique [40] was employed to evaluate the performances of the learning systems described in this paper. The Jack-knife technique is usually implemented to provide a correct statistical evaluation of the performance of a classifier when applied to a limited number of samples divided into two sets: a training set and testing set. In practice, a random number generator is used to select the samples used for training and the samples kept for testing. The classification error varies with the training and testing sample sets and, for a finite number of samples, an error-counting procedure is used to estimate the performance of the classifier [40]. In each experiment described in this paper, 80% of the available samples were randomly selected and used for training while the remaining 20% were used for testing. Therefore 287 samples were used for training and 72 for testing. The data were divided randomly into training and testing sets using the Jack-knife technique. The results were then analyzed to assess the performance.

### 4.2 Optimising Machine learning algorithms

In order to determine which of the two machine learning algorithms is more suitable for predicting DNA binding proteins, SVM and CCNN systems were optimised to find the best parameters and topology, as described in the next two subsections..

### 4.2.1 Optimising the SVM for final comparison

The SVM experiments were carried out using the mySVM program and to optimise performance both the kernel and its parameters were determined empirically because there are no known guidelines to help choose them [41]. For the work reported here, several kernel types were tested including Dot, Polynomial, Radial and ANOVA, The results obtained are illustrated in Figure 2.

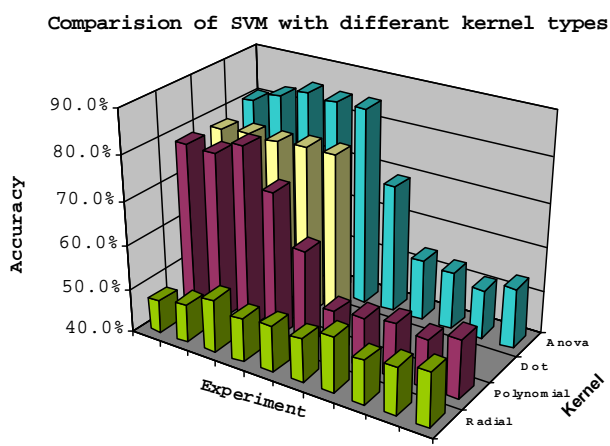

Figure 4. Comparison of SVMs with different kernel types to predict DNA-BPs.

It was found that the ANOVA kernel, which is defined as

$$k(x, y) = \left(\sum \exp(-\gamma(x_i - y_i))\right)^d \tag{9}$$

provides the best results. In equation (9), a sum of products of exponential functions in the x and y directions defines the ANOVA kernel and the shape of the kernel is controlled by the parameters (Gamma) and d (degree).

To complete the SVM optimisation it was necessary to determine the values of these parameters. During each iteration, was incremented from 1 to 100. For each value of gamma values of the seven performance criteria TP, FP, FN, TN, Accuracy, Specificity and Sensitivity were generated, by averaging over ten iterations. The value of d was incremented from 1 to 10. The results for different values of are shown in Figure 5. The best result was obtained using a value of equal to 2.

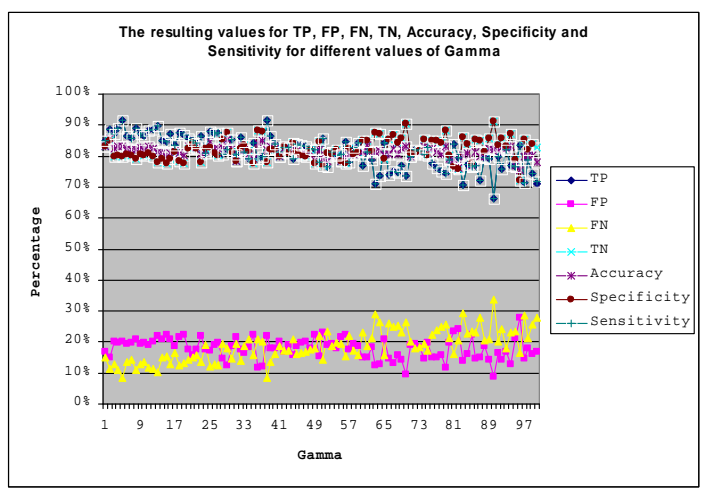

Figure 5. Comparison of SVM performance criteria TP, FP, FN, TN, Accuracy, Specificity and Sensitivity, for different values of

Figure 6 shows the percentage accuracy of correct DNA-BP prediction when the value of is fixed at 2 and the value of d is varied between 1 and 10. As shown in this figure, optimum performance is obtained when the ANOVA kernel used is and the and d parameters are set to 2 and 5 respectively.

Using the ANOVA kernel and these optimised parameters, resulting values obtained were 86.7% for accuracy, 91.1% for sensitivity and 85.3% for specificity.

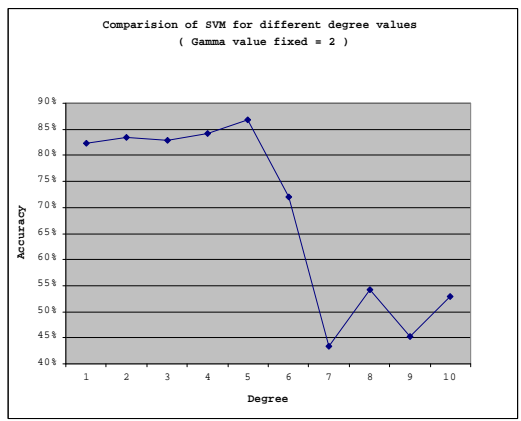

**Figure 6** Percentage accuracy using SVM, with set to 2 and for different values of *d*.

### 4.2.2 Optimising the CCNN for final comparison

The NN has proven to be a very useful tool for solving many real-life problems. However, efficient implementation of the NN usually requires long training sessions [46] which depend on the training vector and on the topology of the NN [47]. Several topologies were investigated in previous studies [30-35][47] for several applications not related to DNA-BP prediction. These topologies included: Elman back-propagation, feed-forward back-propagation and cascade feed forward back-propagation. It was found in all these works that the CCNN provided the best performance in terms of convergence time and optimum topology. Hence, we decided to build on these findings and use a CCNN in our work.

In this network, the first layer has connecting weights to the input layer and each subsequent layer has weights connecting it to all previous layers including the input layer.Two sets of experiments were carried out one using a network with one hidden layer with different value of hidden nodes, and the other using a network with two hidden layers and varying numbers of hidden nodes. From the results obtained it was concluded that a CCNN with 1 hidden layer provided the best results. The network was trained by changing the numbers of nodes from 1 to 50 in the hidden layer. For each number of hidden nodes, the values of the seven performance criteria TP, FP, FN, TN, Accuracy, Specificity and Sensitivity were generated, by averaging over ten iterations. Figure 7 shows the results obtained.

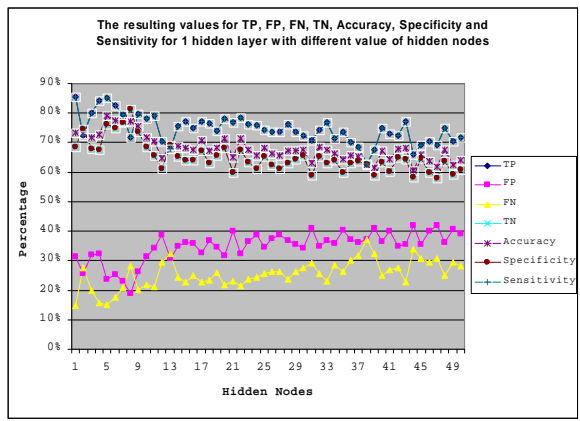

Figure 7 Comparison of performance criteria TP, FP, FN, TN, Accuracy, Specificity and Sensitivity for a CCNN with 1 hidden layer and varying numbers of hidden nodes.

Figure 8 shows the variation of accuracy with number of hidden nodes for a CCNN with 1 hidden layer and it can be seen that 5 hidden nodes gave the best performance. The optimised values obtained for Accuracy, Sensitivity and Specificity were 75.4%, 82.6% and 72.4% respectively.

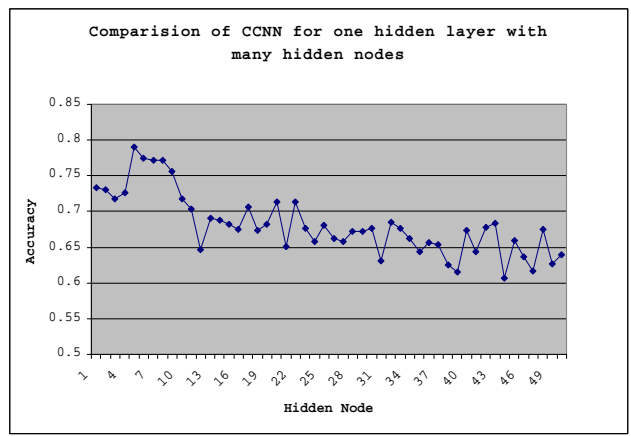

Figure 8 The variation of performance with number of hidden nodes for a CCNN with single hidden layer.

### 4.3. Comparing the performance and result analysis

After optimising SVM and CCNN learning algorithms, we compared their prediction performances. As indicated before the SVM uses the ANOVA kernel with and d parameters set to 2 and 5 respectively and the CCNN has 1 hidden layer with 5 hidden nodes. For all the training and testing experiments, the Jack-knife technique was applied to obtain the training and testing sets. According to the experiments, the CCNN requires longer training times. On the other hand, the SVM was very fast in training and on average provided the best performance, achieving 86.7% successful prediction of DNA-BP. Figure 9 shows a graphical comparison of DNA-BP accuracy for SVM and CCNN.

For this work the learning process for both algorithms continues until a termination criterion is met and the training process has fully converged regardless of the number of training iterations required. We believe that this is a more accurate method of comparing different learning experiments and algorithms than the fixed number of iterations used in [16]. If a fixed number of iterations is used to compare the learning performance then we may end up comparing a fully converged learning system with a partially converged system.

Another difference between our work and [16] lies in the very definition of the Jack-knife method. In [16] the protein sets are randomly separated into N parts. Each time, one part is used for testing and the others are used for training. This is repeated until each part forms the testing dataset exactly once. However, the overall performance depends on the type of proteins within each set. Without this information, it is impossible to re-produce their work. In our work, the training and testing sets are found randomly, and the system accuracy is taken to be the average of 10 learning experiments. This should make it much easier for other researchers to compare their future results with ours.

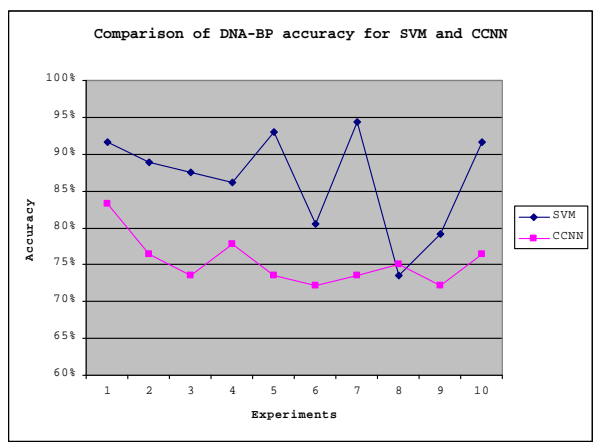

**Figure 9** Graphical comparison of experiments results for DNA-BPs prediction using SVM and CCNN systems.

## 5. CONCLUSIONS AND FUTURE WORK

In this paper, the implementation of two Machine learning systems for the identification of DNA-BPs has been described. The performance of both SVM and CCNN systems were evaluated using the Jack-knife method, and compared to obtain the best prediction results for DNA-BPs. An accuracy of 86.7% for SVM and 75.4 % for CCNN was achieved. These values are comparable with previously published studies giving accuracies ranging from 67% to 100% [11-17]. However, our result obtained using SVM and Jack-knife technique is higher than those reported previously using the same combination.

As explained in the methodology and system design section on feature extraction, selected features used together can distinguish DNA-BPs from non-DNA-binding ones more accurately than when they are used separately [16]. Machine learning systems can exploit this observation. The implementations described here use three classes of features and the results are highly encouraging. It would be straightforward to include new features into our machine learning system, which might improve the performance. Besides including more features, another potential approach for improving the prediction power of the SVM would be through development of models specific to a class, a family or even motifs [49].